# Dynamically evolving segment anything model with continuous learning for medical image segmentation

Zhaori Liu[†], Mengyang Li[†], Hu Han[*], Enli Zhang, Shiguang Shan, Zhiming Zhao[*]

## Affiliations

Institute of Computing Technology, Chinese Academy of Sciences, Beijing, China

Zhaori Liu, Hu Han, Shiguang Shan

Chinese PLA General Hospital, Beijing, China

Mengyang Li, Enli Zhang, Zhiming Zhao

## Contributions

H.H. and Z.Z. conceptualized the research. Z.L. developed the methodology and conducted the dataset evaluation. M.L. and E.Z conducted the clinical evaluation. Z.L., H.H., and M.L. participated in drafting the manuscript, and result analysis. S.S. provided valuable advice and support for the work. All authors revised and approved the manuscript.

## Corresponding author

Correspondence to: Hu Han, and Zhiming Zhao.

Email: hanhu@ict.ac.cn, zhimingzhao616@126.com

## Abstract

Medical image segmentation is essential for clinical diagnosis, surgical planning, and treatment monitoring. Traditional approaches typically strive to tackle all medical image segmentation scenarios via one-time learning. However, in practical applications, the diversity of scenarios and tasks in medical image segmentation continues to expand, necessitating models that can dynamically evolve to meet the demands of various segmentation tasks. Here, we introduce EvoSAM, a dynamically evolving medical image segmentation model that continuously accumulates new knowledge from an ever-expanding array of scenarios and tasks, enhancing its segmentation capabilities. Extensive evaluations on surgical image blood vessel segmentation and multi-site prostate MRI segmentation demonstrate that EvoSAM not only improves segmentation accuracy but also mitigates catastrophic forgetting. Further experiments conducted by surgical clinicians on blood vessel segmentation confirm that EvoSAM enhances segmentation efficiency based on user prompts, highlighting its potential as a promising tool for clinical applications.

## Introduction

Medical image segmentation plays a crucial role in clinical diagnosis[1], surgical planning[2], and treatment monitoring[3]. Despite its significance, numerous clinical applications still rely on entirely manual segmentation. This approach demands substantial expertise and is extremely time-consuming. This reliance presents significant challenges in segmenting large volumes of data. With the advancement of deep learning powered by large datasets and high computational resources, foundational AI models have exhibited capabilities nearing or even surpassing human expertise across various tasks[4-10]. The recent Segment Anything Model (SAM)[9], pre-trained with over 1 billion masks, has gained attention as a versatile promptable segmentation model designed to "segment anything", such as humans, vehicles, animals, plants, etc, significantly improving image segmentation efficiency. The medical adaptation, MedSAM[11], was developed through extensive training on over 1.5 million masks derived from publicly accessible medical data, with the goal of transforming SAM into a domain-specific expert in medical imaging. MedSAM enables the segmentation of various objects in different modalities of medical images (CT, MRI, surgical image, etc.) by simply requiring a box prompt around the region of interest, significantly simplifying the segmentation process for complex medical data.

Despite the advantages of SAM and MedSAM, both models face limitations stemming from their reliance on one-time training or fine-tuning on large-scale static datasets. Medical data, by contrast, often evolves variations dynamically, due to changing clinical demands, variations in data sources, advancements in imaging technology, etc[12]. Therefore, even foundation models like SAM or MedSAM may experience difficulties in obtaining high segmentation performance in dynamically evolving scenarios. In contrast, humans can continuously accumulate new knowledge from new experiences. This mechanism of human learning has inspired the development of continual learning methods[13-16], and there have been some continual learning approaches designed for convolutional neural network (CNN)[17] based medical image segmentation, e.g. lifelong nnU-Net[18] and MoCSS[19]. However, existing continual learning methodologies, when applied to foundational segmentation models such as SAM, may encounter issues of overfitting and catastrophic forgetting. This vulnerability arises from the substantial parameter capacity inherent in these models, combined with the limitations in dataset size encountered during continual learning tasks[20]. The dynamic evolution of foundational segmentation models through continuous learning remains an underexplored area of research.

Based on this insight, we propose EvoSAM, a dynamically evolving model for medical image segmentation. EvoSAM utilizes lightweight task-specific Low-Rank Adaptation (LoRA)[21] experts to allow SAM to accumulate new knowledge from evolving scenarios, thereby progressively enhancing its segmentation capabilities. Each LoRA expert is independently learned on distinct continual learning tasks. For inference, the most appropriate LoRA expert is selected via a ridge regression-based[22] expert matcher, ensuring SAM applies the most relevant specialist knowledge to a given test image.

We evaluate our method in two representative continual learning scenarios (see Fig. 1) within medical imaging: (1) blood vessel segmentation in surgical images with continuously expanding blood vessel categories and (2) multi-site prostate MRI segmentation[23] across various medical centers. We also benchmark our approach against existing continual learning methods[24-26] applied to SAM. The experimental results demonstrate that our method effectively mitigates, and in some instances nearly eliminates, catastrophic forgetting, allowing SAM or MedSAM to progressively expand its segmentation capabilities. In addition, when our model is employed to assist surgeons in segmenting blood vessels in surgical images, it substantially boosts segmentation efficiency, as reflected by improved initial Dice scores and decreased segmentation time per image.

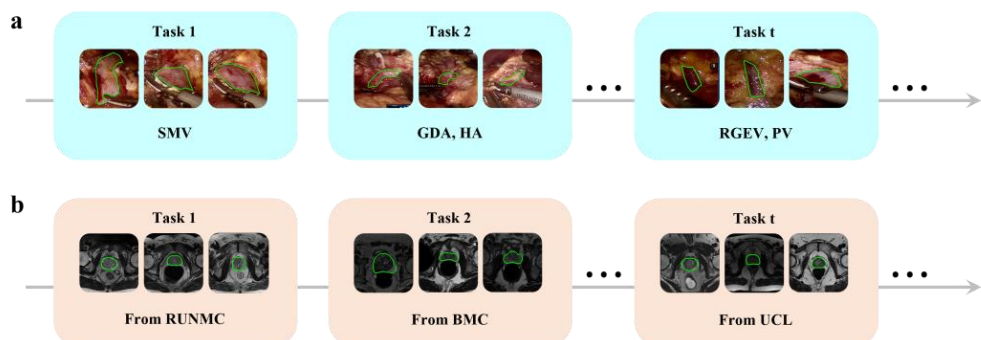

**Fig. 1: Two representative medical image segmentation scenarios require continuous learning to dynamically evolve the model**: (a) blood vessel segmentation in surgical images with continuously expanding blood vessel categories, and (b) prostate MRI segmentation across various medical centers.

## Results

We evaluate the effectiveness of EvoSAM under two continual learning scenarios in medical image segmentation: blood vessel segmentation in surgical images with continuously expanding blood vessel categories and prostate MRI segmentation across various medical centers. For blood vessel segmentation in surgical images, we consider two experimental protocols with 3 tasks and 5 tasks, respectively. In each protocol, different tasks include distinct sets of blood vessel categories, simulating a situation where the blood vessel categories to be segmented increase progressively. For multi-site prostate MRI segmentation[23], we consider the protocol with 6 tasks, each corresponding to a different clinical site with varying prostate MRI imaging conditions. Supplementary Table 3 details the tasks of each scenario.

To compare with our EvoSAM, we also apply sequential fine-tuning (Seq FT), experience replay (ER)[24], distillation[25], and elastic weight consolidation (EWC)[26] to SAM. Seq FT is a common approach allowing a pre-trained model to learn from tasks one at a time. ER is a representative replay-based continual learning method; distillation and EWC are popular regularization-based continual learning methods.

Additionally, we combined the training data from all tasks to fine-tune SAM (referred to as SAM-Joint), which serves as an upper bound performance of continual learning methods.

We refer to existing continual learning literature[27-29] and introduce two metrics to quantitatively evaluate the performance of our model: average mDice and average forgetting. These metrics respectively measure the average segmentation performance across all tasks and the degree of performance degradation of early tasks.

## EvoSAM Outperforms Representative Continual Learning Strategies

**Blood vessel segmentation in surgical images.** We begin by assessing the efficacy of various methods for blood vessel segmentation in surgical imagery. In practical applications, the sequence in which tasks are encountered is often unpredictable, meaning that a method effective for one specific task sequence may not perform well for another. To mitigate the potential bias introduced by a singular task sequence, we conduct multiple continual learning experiments under both a 3-task and a 5-task protocol. Specifically, for the 3-task protocol, we evaluate all six permutations of the task sequences, whereas for the 5-task protocol, we select six sequences from all possible permutations. Within each sequence, we calculate the average mDice and average forgetting metrics across all tasks and aggregate these metrics across all sequences.

Tables 1a and 1b present the segmentation performance of our EvoSAM framework against several baseline continual learning strategies applied to SAM. The results indicate that direct application of SAM for blood vessel segmentation leads to poor average mDice scores in both the 3-task and 5-task protocols. This is expected due to the considerable divergence between SAM's original training dataset and the surgical image dataset. The straightforward fine-tuning approach, Seq FT, significantly enhances blood vessel segmentation performance, achieving average mDice scores of 58.26 vs. 52.23 in the 5-task protocol and 64.87 vs. 51.43 in the 3-task protocol. However, Seq FT exhibits notable catastrophic forgetting for early tasks, with an average forgetting of 3.87% and 5.73% in the 5-task and 3-task protocols. This is because Seq FT updates the parameters of the full SAM model. Among the three representative continual learning baselines, EWC does not consistently improve average mDice or mitigate average forgetting. Under the 3-task protocol, EWC improves the average mDice by 0.59% and reduces the average forgetting by 0.72% relative to Seq FT; however, under the 5-task protocol, it improves the average mDice by 1.22% but also raises average forgetting by 0.25%. Distill and ER (25%) outperform Seq FT by improving average mDice by at least 4.63% and reducing average forgetting by at least 1.53%. Notably, Distill achieves negative average forgetting (-0.07%) under the 3-task protocol, suggesting that this strategy can sometimes improve the blood vessel segmentation accuracy for earlier tasks while learning new knowledge. It is worth mentioning that ER requires storing data from previous tasks, which might be impractical in privacy-sensitive or storage-constrained scenarios. Compared to existing methods, our EvoSAM achieves promising results in both average mDice (72.42% and 74.34% under the 5-task and 3-task protocols, respectively) and average forgetting (0.04% and 0.07% under the 5-task and 3-task protocols, respectively). Moreover, EvoSAM approaches the performance of the upper bound method SAM-Joint.

We also apply the proposed continual learning approach and baseline continual learning approaches to MedSAM to perform blood vessel segmentation and report the results in Tables 1c and 1d. Similar conclusions can be drawn under the 5-task and 3-task blood vessel segmentation protocols. Notably, EvoSAM can effectively mitigate the gap between natural images and medical images, achieving continual learning performance comparable to EvoMedSAM.

**Multi-site prostate MRI segmentation.** Table 1e presents the multi-site prostate segmentation performance of our EvoSAM model against several baseline continual learning strategies applied to SAM. Since MedSAM has used the multi-site prostate segmentation dataset for training, we only use SAM for continual learning while directly report the result by MedSAM for comparison. The results indicate that direct application of SAM for multi-site prostate segmentation leads to a poor average mDice score of 85.15. Seq FT improves the average mDice score to 91.60, but it still has a relatively large average forgetting of 0.75%. EWC decreases the average mDice by 1.84% and increases average forgetting by 0.36% compared to Seq FT. Distill and ER (25%) demonstrate effective anti-forgetting capabilities and better overall performance. Under this protocol, EvoSAM exhibited nearly zero forgetting, with average mDice exceeding the best of the three traditional continual learning strategies by 0.38%, outperforming MedSAM by 1.0%, and achieving results comparable to the upper bound performance by SAM-Joint.

**Table 1 Average mDice and average forgetting of blood vessel segmentation and multi-site prostate MRI segmentation by our method and baseline methods for continual learning. a** Blood vessel segmentation with SAM-based continual learning in 5-task protocol. **b** Blood vessel segmentation with SAM-based continual learning in 3-task protocol. **c** Blood vessel segmentation with MedSAM-based continual learning in 5-task protocol. **d** Blood vessel segmentation with MedSAM-based continual learning in 3-task protocol. **e** Multi-site prostate MRI segmentation with SAM-based continual learning. ER (25%) indicates that 25% of the data from each task is randomly selected for replay. EvoMedSAM is a variant of EvoSAM that is applied to the MedSAM.

| a | | | b | | |
|---|---|---|---|---|---|
| **Method** | **Average mDice** | **Average forgetting** | **Method** | **Average mDice** | **Average forgetting** |
| SAM | 52.23 | N/A | SAM | 51.43 | N/A |
| SAM-Seq FT | 58.26±3.77 | 3.87±1.41 | SAM-Seq FT | 64.87±2.83 | 5.73±1.56 |
| SAM-EWC | 59.48±3.20 | 4.12±0.63 | SAM-EWC | 65.46±3.72 | 5.01±1.89 |
| SAM-Distill | 63.00±1.81 | 2.34±0.90 | SAM-Distill | 70.10±1.23 | **-0.07±1.04** |
| SAM-ER (25%) | 67.45±1.80 | 0.63±0.66 | SAM-ER (25%) | 69.50±0.71 | 0.48±0.83 |
| **EvoSAM** | **72.42±1.08** | **0.04±0.23** | **EvoSAM** | **74.34±1.00** | 0.07±0.23 |
| SAM-Joint | 75.69 | N/A | SAM-Joint | 76.27 | N/A |

| c | | | d | | |
|---|---|---|---|---|---|
| **Method** | **Average mDice** | **Average forgetting** | **Method** | **Average mDice** | **Average forgetting** |
| MedSAM | 68.76 | N/A | MedSAM | 69.00 | **N/A** |
| MedSAM-Seq FT | 63.24±3.90 | 4.20±1.90 | MedSAM-Seq FT | 68.32±2.08 | 4.75±1.76 |
| MedSAM-EWC | 61.97±4.08 | 5.12±1.20 | MedSAM-EWC | 68.30±1.56 | 4.65±1.70 |
| MedSAM-Distill | 70.15±2.12 | 0.96±0.78 | MedSAM-Distill | 72.51±0.78 | 0.58±1.14 |
| MedSAM-ER (25%) | 70.14±0.76 | 0.66±0.44 | MedSAM-ER (25%) | 71.17±0.97 | 1.65±1.13 |
| **EvoMedSAM** | **73.49±0.78** | **0.14±0.19** | **EvoMedSAM** | **75.06±0.74** | **0.11±0.14** |
| MedSAM-Joint | 76.04 | N/A | MedSAM-Joint | 76.27 | N/A |

| e | | |
|---|---|---|
| Method | Average mDice | Average forgetting |
| SAM | 85.15 | N/A |
| MedSAM | 92.33 | N/A |
| SAM-Seq FT | 91.60±0.80 | 0.75±0.18 |
| SAM-EWC | 89.76±2.51 | 1.11±0.60 |
| SAM-Distill | 92.60±0.35 | **-0.06±0.16** |
| SAM-ER(25%) | 92.95±0.23 | 0.07±0.24 |
| **EvoSAM** | **93.33±0.19** | 0.00±0.01 |
| SAM-Joint | 93.55 | N/A |

## EvoSAM Demonstrates Continuous Improvement in Learning New Knowledge with Minimal Forgetting

We analyze the performance variations of our method and baseline methods throughout the continual learning process of SAM under the 5-task blood vessel segmentation protocol. Figure 2 presents the average mDice scores for all five tasks' testing sets across different stages of continual learning. Given that EWC did not exhibit substantial performance improvements in this specific experimental setting (as shown in Table 1), we have opted not to include its results in the subsequent analysis. Occasionally, Seq FT exhibits continuous improvement in average mDice for a certain task sequence (as shown in Fig. 2b). However, in most cases, it shows significant performance fluctuations and ultimately fails to achieve notable gains by the final task (as depicted in Fig. 2a, 2c, 2d, 2e, 2f). This can be attributed to the progressive overfitting[19] (further discussed in Supplementary Results) and catastrophic forgetting of SAM. By contrast, Distill and ER (25%) achieve higher overall average mDice scores in the continual learning process, owing to their partial mitigation of forgetting (as shown in Table 1 and Fig. 4). Nevertheless, their learning curves still exhibit considerable fluctuations (as shown in Fig. 2b, 2c, 2d, 2e, 2f). In contrast, EvoSAM demonstrates robust and stable improvements across different task sequences, significantly outperforming baseline methods (Seq FT, Distill, ER (25%)). Its consistent performance gains highlight its ability to effectively mitigate forgetting and maintain a balance between acquiring new knowledge and retaining prior task information.

We also analyze the performance variations of our method and baseline methods throughout the continual learning process of MedSAM under the 5-task blood vessel segmentation protocol (as shown in Fig. 3). Under this experimental setting, Seq FT exhibits more severe performance degradation, with final results that are even worse than those prior to continual learning (MedSAM). While Distill and ER (25%) show improvements over Seq FT, they fail to derive significant benefits from the continual learning process. By contrast, the proposed EvoMedSAM once again achieves stable and consistent performance improvements, highlighting its effectiveness in addressing the challenges of continual learning.

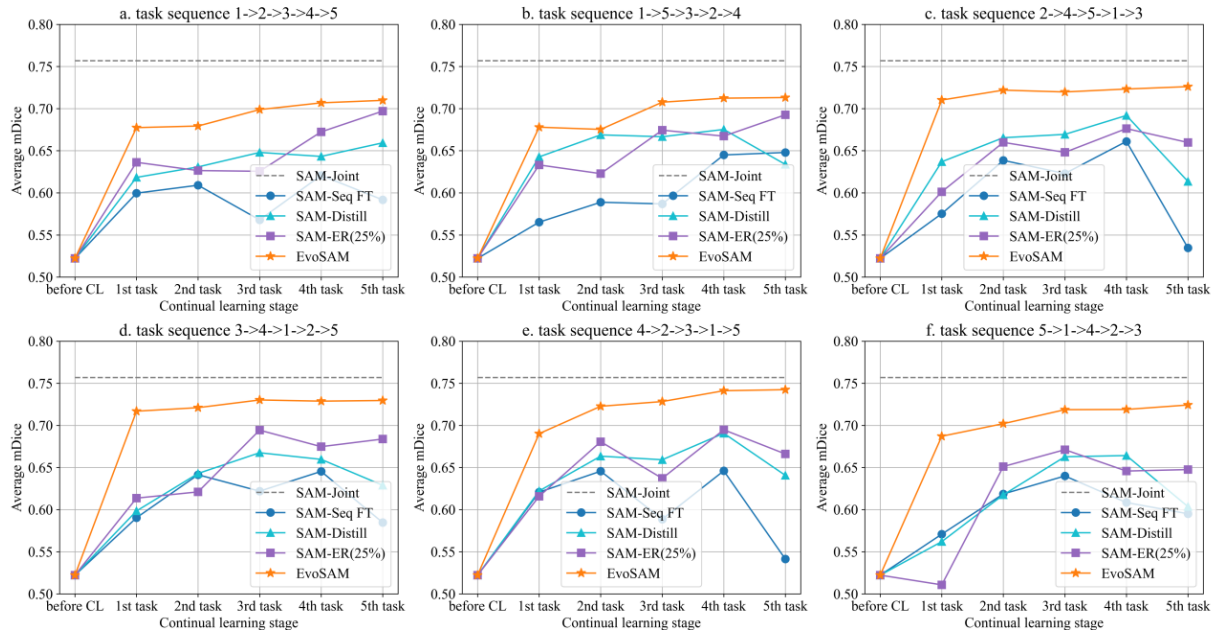

**Fig. 2: Average mDice of blood vessel segmentation in the 5-task protocol by our method and baseline methods for continual learning with SAM under six different task sequences.**

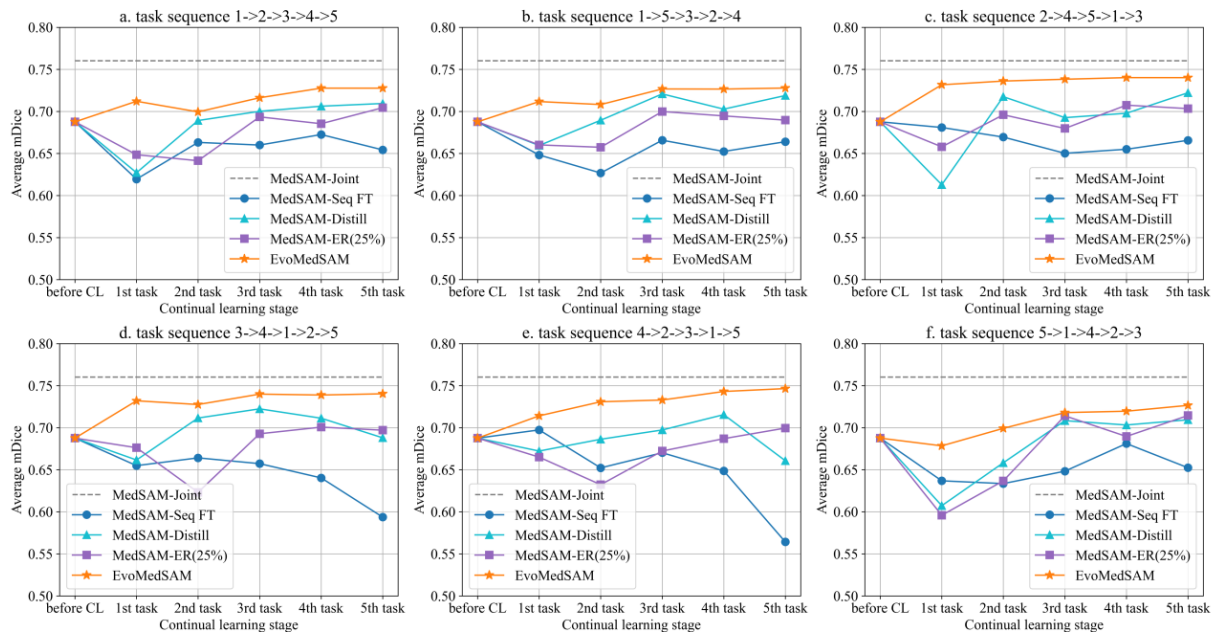

**Fig. 3: Average mDice of blood vessel segmentation in the 5-task protocol by our method and baseline methods for continual learning with MedSAM under six different task sequences.**

We conducted an in-depth analysis of performance variations, comparing our method with baseline techniques during the continual learning process of SAM, under the 5-task blood vessel segmentation protocol, as evaluated on each task's test set. Figure 4 presents the mDice scores for individual methods in one specific task sequence, namely 4→2→3→1→5, depicted in Fig. 2e. Within each matrix, the diagonal elements represent the capability of acquiring knowledge from new tasks, and the elements in the lower triangular region indicate the capability of preserving knowledge from prior tasks. Among the four matrices, the diagonal elements in Fig. 4a are notably small, suggesting that Seq FT cannot consistently acquire knowledge from new tasks. In the lower triangular matrix, almost every column shows an overall decreasing trend, highlighting Seq FT's severe forgetting issue. Although Distill and ER exhibit some advancements in both acquiring new knowledge and retaining prior knowledge, these improvements are relatively minor (Fig. 4b and Fig. 4c). In contrast, EvoSAM demonstrates a significant improvement in both acquiring new task knowledge and retaining prior task knowledge, as illustrated in Fig. 4d. Notably, the columns in the lower triangular matrix do not present a severe declining trend, indicating effective mitigation of catastrophic forgetting. Interestingly, the mDice score for 2nd task progressively improves as subsequent tasks are learned. This suggests that, unlike the typical catastrophic forgetting associated with continual learning, EvoSAM has the potential to improve earlier task performance through the accumulation of knowledge from new tasks.

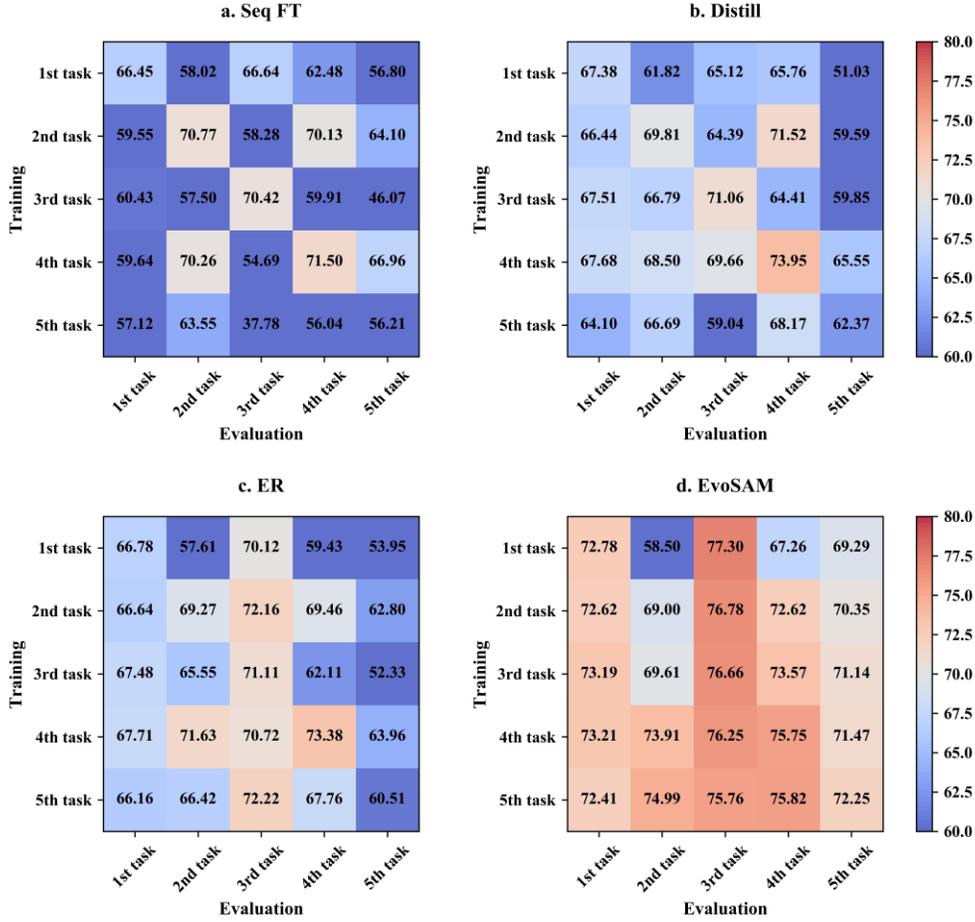

**Fig. 4: The mDice matrices for different continual learning models based on SAM are shown under the 5-task blood vessel segmentation protocol, following the task order: 4→2→3→1→5, labeled as the 1st to 5th tasks, respectively.** The value in the i-th row and j-th column represents the mDice of the model on the j-th task's test set after continual learning with the 1st to the i-th task's data.

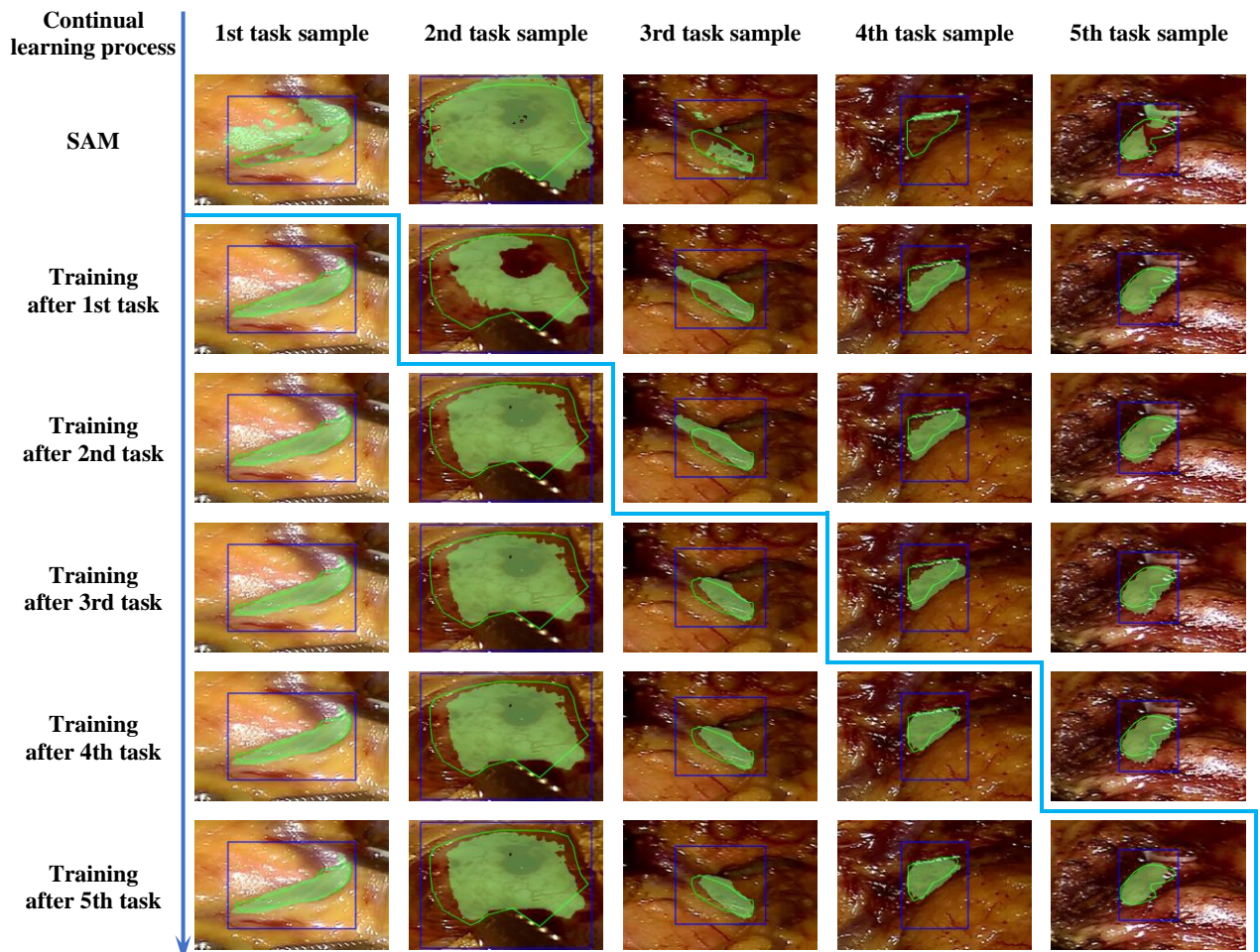

**Fig. 5: Visualization of blood vessel segmentation by EvoSAM after learning each task during continual learning.** The segmentation results after EvoSAM learns each task are shown for five surgical images across five tasks with a task sequence of 4(1st)→2(2nd)→3(3rd)→1(4th)→5(5th)) under the 5-task protocol.

To better illustrate the effectiveness of EvoSAM during continual learning, we present its segmentation results for five surgical images across five tasks with a task sequence of 4→2→3→1→5 under the 5-task protocol. As illustrated in Fig. 5, before continual learning, SAM exhibits relatively satisfactory results on the image of the 2nd task, though with irregular and uneven boundaries. Conversely, its segmentation results for the images of the 1st, 3rd, and 5th tasks are dispersed and imprecise, with only a minor portion of the blood vessels being accurately segmented in the 4th task. After learning on the 1st task, EvoSAM demonstrates significantly enhanced segmentation precision for the corresponding surgical images. We also observe noticeable improvements in the segmentation of surgical images from unseen tasks, specifically the 3rd, 4th, and 5th tasks. An exception arises with the 2nd task, where the segmented regions show significant shrinkage, and a substantial portion of the blood vessel area is erroneously classified as background. However, after learning on the 2nd task, the segmentation performance on its image improves considerably, capturing more blood vessel areas with relatively smooth boundaries, while maintaining robust and accurate segmentation results for the 1st task. This trend continues consistently throughout the subsequent phases of continual learning. These results further demonstrate the capability of our EvoSAM system to effectively accumulate new knowledge while minimizing the forgetting of previously acquired knowledge.

**EvoSAM Enables Faster Segmentation for Surgical Clinicians**

To evaluate the practical effectiveness of EvoSAM, we conducted an experiment in which surgical clinicians of different years of experience (associate chief surgeon, resident surgeon, trainee) performed blood vessel segmentation using four different models: SAM, MedSAM, EvoSAM, and EvoMedSAM. We developed a blood vessel segmentation program (refer to Supplementary Fig. 40-43) using PyQt5 running on a laptop workstation with RTX 3070 GPU, which allows each clinician to segment each blood vessel instance by using all four models in anonymous order. We chose 100 blood vessel instances from the 9 blood vessel categories contained in the 5-task blood vessel segmentation protocol. EvoSAM and EvoMedSAM models are trained throughout all 5 tasks.

Surgical clinicians began by selecting the region of interest (ROI) containing the target blood vessel in the image with a box prompt. The model then generated an initial segmentation result, which clinicians could refine using brush and eraser tools until a satisfactory result was achieved. Since the SAM encoder takes a long time to extract image features, we pre-compute the image feature offline. To compare the performance of the models before and after continual learning, we recorded the dice score for each blood vessel instance and the time spent by clinicians in manually refining the initial segmentation results. Due to varying levels of experience among clinicians, the segmentation accuracy achieved when each clinician was satisfied may differ significantly. Therefore, we used the segmentation results from the most experienced clinician as the ground truth (confirmed by the chief surgeon) and only considered instances where the IoU of the clinician's final segmentation result with the ground truth was $\geq$ 0.7. We excluded cases with extreme segmentation times ($\geq$ 40sec.) and evaluated the models based on the initial segmentation results and the time spent on manual correction for the remaining instances.

**a**

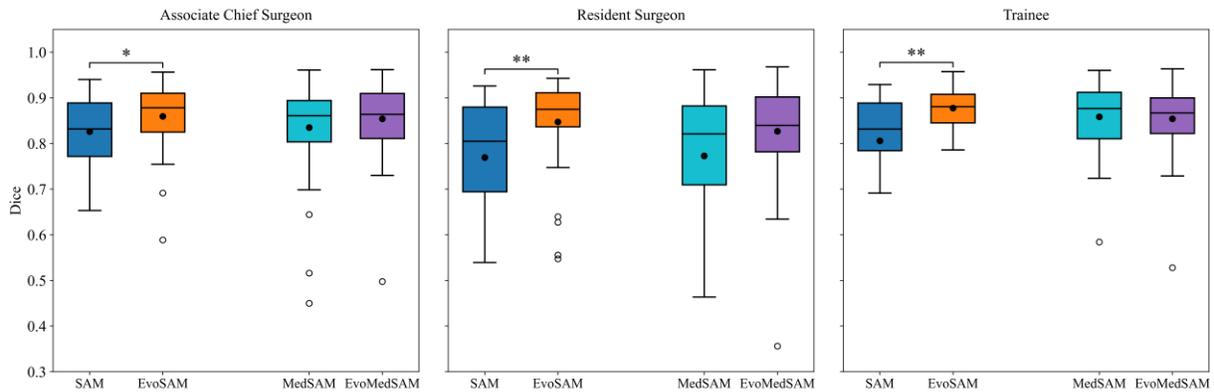

**b**

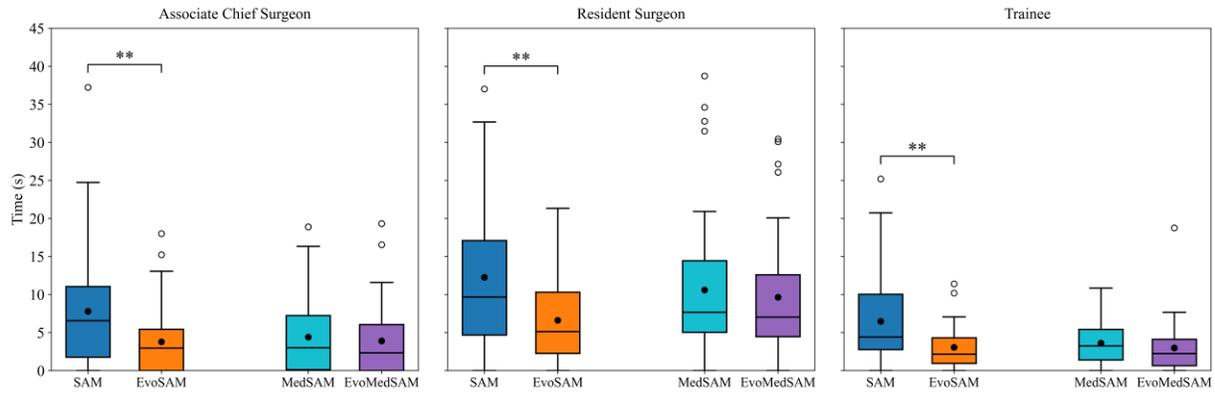

**Fig. 6: Comparisons between different segmentation models when performing semi-automatic blood vessel segmentation by three surgical clinicians. a** The dice scores of the initial segmentation by different models given a user-provided bounding box for each blood vessel. **b** The time (in Sec.) spent on refining the initial segmentation result until a satisfied result is achieved. The significance test was performed using the Wilcoxon rank-sum test.

Fig. 6a presents the dice scores of the initial segmentation results generated by each model, while Fig. 6b shows the time surgical clinicians spent on manual refinement. The results demonstrate that SAM frequently produced suboptimal initial segmentations, which required extensive manual refinement. In contrast, the EvoSAM model demonstrated an impressive improvement in initial segmentation accuracy, significantly reducing the need for manual adjustments. The MedSAM model, benefiting from pre-training on large-scale medical datasets, sometimes achieved more accurate initial segmentations than SAM, thus decreasing the necessity for manual refinement. Similar to EvoSAM, EvoMedSAM not only improved the initial segmentation accuracy (except for the trainee clinician) but also reduced the manual refinement time.

## Discussion

In this study, we propose EvoSAM, a dynamically evolving medical image segmentation model, to address the challenge of medical image segmentation under dynamically evolving scenarios. EvoSAM employs LoRA experts to enable SAM to continuously expand its segmentation capability by accumulating new knowledge from dynamically evolving scenarios. It utilizes a ridge regression-based expert matcher to handle progressively increasing tasks, enabling flexible matching of task-specific parameters. We evaluated our method on progressively expanding blood vessel segmentation tasks and multi-site prostate segmentation tasks, with various task orders. The experimental results indicate that EvoSAM can acquire new knowledge continuously while preserving prior knowledge, outperforming several representative continual learning strategies in terms of average mDice. Furthermore, clinical studies by surgical clinicians using different segmentation models showed that EvoSAM can improve annotation efficiency.

Despite its promising continual learning capabilities, EvoSAM has certain limitations. One limitation is its inability to model shared knowledge across tasks, which could hinder the model's ability to learn new knowledge when individual tasks have limited data. Furthermore, EvoSAM exhibits slight parameter growth, which, although having a negligible impact on inference time, might decrease efficiency in resource-constrained environments when the number of tasks becomes huge.

Future research could focus on enhancing EvoSAM and other SAM-based continual learning models to enable efficient application in highly constrained scenarios, such as online learning. This would contribute to more flexible continual learning frameworks for SAM. Additionally, applying EvoSAM to personalized continual learning for individual clinicians is an intriguing avenue, where incorporating human feedback-based learning could further enhance its practical utility. Overall, EvoSAM is an effective and reliable SAM-based continual learning model, and we look forward to seeing its potential impact across a variety of continual learning scenarios in medical applications.

## Methods

### Dynamically Evolving the Segment Anything Model

Segment Anything Model (SAM) is a foundational segmentation model with a Transformer-based architecture[30]. It contains three main components: an image encoder, a prompt encoder, and a mask decoder. The image encoder employs a pre-trained Vision Transformer (ViT)[31,32] to generate embeddings for an input image. It incorporates window-based attention[33] to handle the high-resolution images, reducing computational cost and inference time. The prompt encoder processes user-provided prompts, such as points, boxes, text, or masks, and converts them into prompt embeddings. The mask decoder takes the image embeddings and prompt embeddings as inputs to produce the segmentation mask. This module consists of a variant of a Transformer decoder block, which incorporates both self-attention and cross-attention mechanisms, and a dynamic mask prediction head that generates the final segmentation result. The SAM model contains a substantial number of parameters, making it prone to overfitting when fine-tuned on small to medium-sized datasets. Recently, parameter-efficient tuning methods[21,34-41] has garnered widespread interest, which adjusts only a subset of model parameters or introduces additional lightweight parameters. Our experimental results (as shown in Supplementary Fig 36) demonstrate that a representative parameter-efficient method, LoRA, significantly improves SAM's performance in certain medical adaptation scenarios.

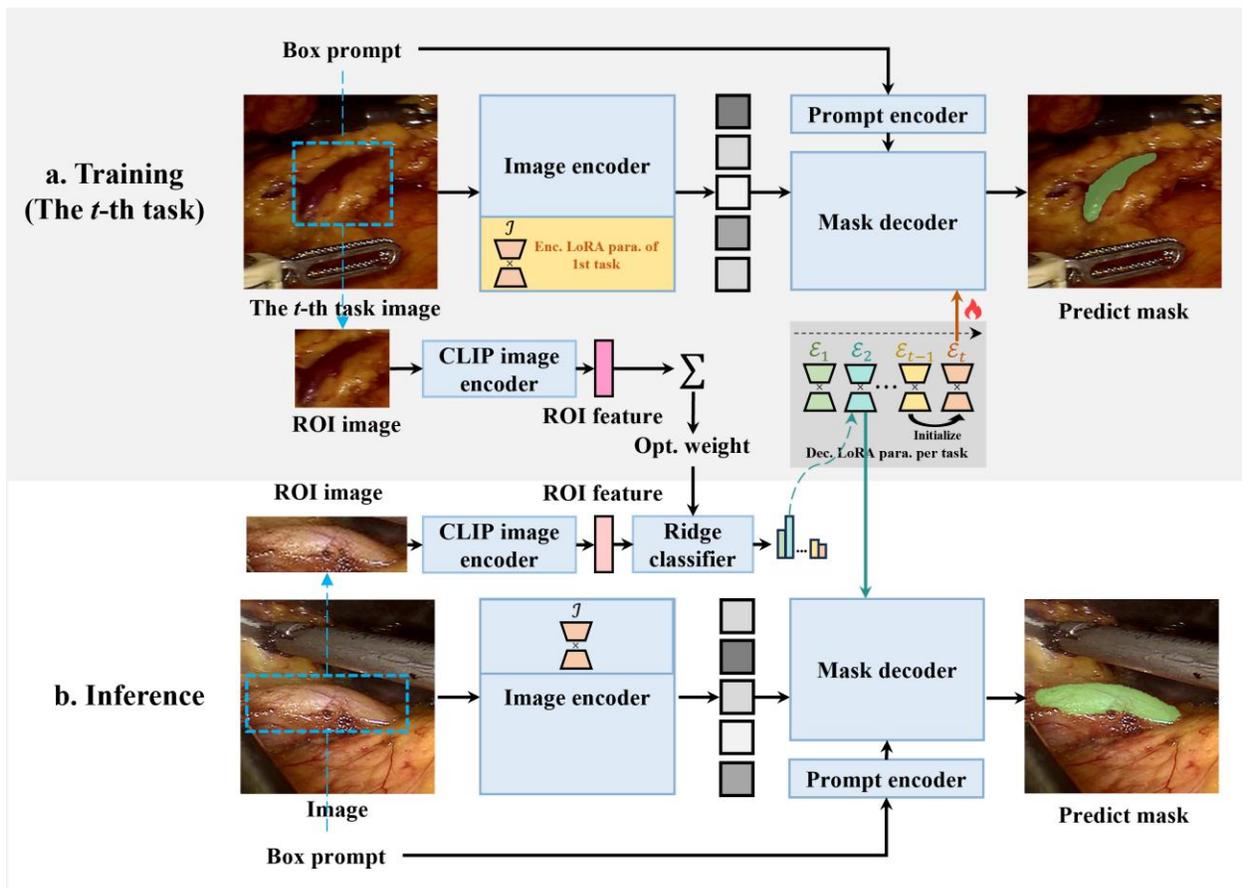

**Fig. 7: Illustration of the architecture of EvoSAM. a** In the training phase, EvoSAM retains all of SAM's pre-trained parameters and independently learns a distinct set of LoRA parameters ($\mathcal{E}_i$) for the mask decoder specific to each task, as well as a set of LoRA parameters ($\mathcal{J}$) for the image encoder for the initial task. Additionally, EvoSAM develops a ridge regression-based expert matcher to select the optimal decoder LoRA parameters for image inference, guided by the user's box prompt (ROI). **b** During the inference phase, the frozen encoder parameters, together with the encoder LoRA parameters, are utilized to extract image features. Concurrently, the frozen decoder parameters and the selected optimal decoder LoRA parameters are employed to generate the segmentation masks.

EvoSAM extends SAM by incorporating a continual learning approach based on LoRA parameter isolation. As shown in Fig. 7, in the training process, it learns a separate set of LoRA parameters for the mask decoder tailored to each task, while also establishing a set of LoRA parameters for the image encoder for the initial task. EvoSAM also learns a ridge regression-based expert matcher to select the most appropriate set of LoRA parameters for any given test image during inference.

In EvoSAM, LoRA parameters for the mask decoder are incorporated into the attention layer (see Fig. 8). Let $Q_{in}, K_{in}, V_{in} \in \mathbb{R}^{n \times d}$ represent the inputs to the attention layer, and the attention is computed as:

$$\text{Attention}(Q_{in}, K_{in}, V_{in}) = \text{softmax}\left(\frac{QK^T}{\sqrt{d}}\right)VW^{(O)}, \tag{1}$$

where $Q, K, V$ are given by:

$$Q = Q_{in}W^{(Q)}, \qquad K = K_{in}W^{(K)}, \qquad V = V_{in}W^{(V)}, \tag{2}$$

where $W^{(Q)}, W^{(K)}, W^{(V)}, W^{(O)} \in \mathbb{R}^{d \times d}$ are the pretrained parameters, and we freeze all these parameters when we introduce the additional trainable LoRA parameters $A^{(Q)}, A^{(V)} \in \mathbb{R}^{d \times r}$ and $B^{(Q)}, B^{(V)} \in \mathbb{R}^{r \times d}$. Unless otherwise specified, $A^{(Q)}$ and $A^{(V)}$ are initialized with random values sampled from a normal distribution, while $B^{(Q)}$ and $B^{(V)}$ are initialized to zero. During inference, the frozen parameters and the newly learned LoRA parameters are used together:

$$W_{new}^{(Q)} = W^{(Q)} + A^{(Q)}B^{(Q)}, \tag{3}$$

$$W_{new}^{(V)} = W^{(V)} + A^{(V)}B^{(V)}, \tag{4}$$

Typically, the rank $r$ is chosen to be much smaller than $d$, resulting in significantly fewer trainable parameters compared to $W^{(Q)}$ and $W^{(V)}$. This reduction not only improves computational efficiency but also enhances the robustness of the adapted model.

We define the training dataset of the $i$-th task in continual learning as $D_i = \{(x_{i,j}, p_{i,j}, m_{i,j})\}_{j=1}^{N_i}$, where $x_{i,j}, p_{i,j}$ and $m_{i,j}$ denote the image, bounding-box prompt, and ground-truth mask of the $j$-th sample in the $i$-th task, respectively. Let $f_{enc}, f_{pr}$ and $f_{dec}$ represent the image encoder, prompt encoder, and mask decoder of SAM, respectively. For the $i$-th task, a separate set of LoRA parameters is introduced into the attention layer of the mask decoder:

$$\mathcal{E}_i = \left\{ A_{d_i,l}^{(Q)}, B_{d_i,l}^{(Q)}, A_{d_i,l}^{(V)}, B_{d_i,l}^{(V)} \mid l = 1, 2, \ldots, L_{dec} \right\} \tag{5}$$

Here, $A_{d_i,l}^{(Q)}, B_{d_i,l}^{(Q)}, A_{d_i,l}^{(V)}, B_{d_i,l}^{(V)}$ are the LoRA parameters in the $l$-th attention layer of the mask decoder. Similarly, a set of LoRA parameters is introduced into the image encoder for the first task:

$$\mathcal{I} = \left\{ A_{e,l}^{(Q)}, B_{e,l}^{(Q)}, A_{e,l}^{(V)}, B_{e,l}^{(V)} \mid l = 1, 2, \ldots, L_{enc} \right\} \tag{6}$$

Here, $A_{e,l}^{(Q)}, B_{e,l}^{(Q)}$ and $A_{e,l}^{(V)}, B_{e,l}^{(V)}$ are the LoRA parameters in the $l$-th attention layer of the image decoder.

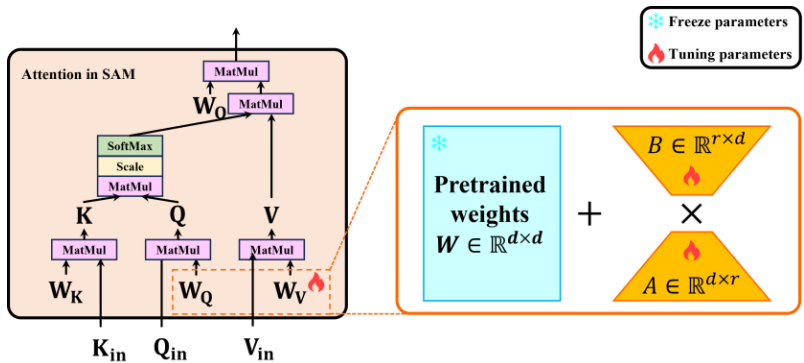

**Fig. 8: Trainable LoRA parameters for the attention layer in EvoSAM.** EvoSAM introduces trainable LoRA parameters $A \in \mathbb{R}^{d \times r}$ and $B \in \mathbb{R}^{r \times d}$ for the projection matrices $W_Q$ and $W_V$ in the attention layer of either the image encoder or the mask decoder.

Following MedSAM[11], EvoSAM uses the sum of the dice loss[42] and binary cross-entropy (BCE) loss to optimize the LoRA parameters for the image encoder and mask decoder:

$$\mathcal{L} = \mathcal{L}_{Dice} + \mathcal{L}_{BCE} \tag{7}$$

When the first task data $D_1$ arrives, the LoRA parameters of the image encoder and the first task LoRA parameters of the mask decoder are jointly optimized using backpropagation[43] and gradient descent:

$$\min_{\mathcal{I},\mathcal{E}_1} \frac{1}{N_1} \sum_{j=1}^{N_1} \mathcal{L}\big(f_{dec}(f_{enc}(x_{1,j}; \mathcal{I}), f_{pr}(p_{1,j}); \mathcal{E}_1), m_{1,j}\big) \tag{8}$$

When the subsequent tasks' data $D_i (i \geq 2)$ arrives, the image encoder LoRA parameters $\mathcal{I}$ are frozen, and only the $i$-th task LoRA parameters $\mathcal{E}_i$ are optimized:

$$\min_{\mathcal{E}_i} \frac{1}{N_i} \sum_{j=1}^{N_i} \mathcal{L}\big(f_{dec}(f_{enc}(x_{i,j}; \mathcal{I}), f_{pr}(p_{i,j}); \mathcal{E}_i), m_{i,j}\big) \tag{9}$$

We initialize the $i$-th task LoRA parameters of the mask decoder from the preceding task.

## LoRA Expert Matcher

After distinct sets of LoRA parameters are learned for individual tasks, EvoSAM needs to identify the most appropriate set of LoRA parameters for an input testing image. To achieve this, we use the pre-trained CLIP[8] image encoder to extract features for each region of interest (ROI) specified by the user, and use a ridge regression-based LoRA expert matcher to determine which set of LoRA parameters should be used for a test image during inference(Fig. 8b). The LoRA expert matcher learning is inherently a class-incremental problem. In this scenario, the number of classes to be classified increases as the continual learning process proceeds, and data from previous tasks is also unavailable when learning from a new task. Inspired by RanPAC[44], a state-of-the-art method for class-incremental learning, we employ an incremental ridge regression classifier to address this challenge.

Formally, let $y_{i,j}$ represent the one-hot vector of the $j$-th training sample's class label (either task-level or task-specific subclass) for the $i$-th task. Denote $N_i$ as the number of samples in $i$-th task, $f_{CLIP}$ as the CLIP image encoder, and $\text{crop}$ as the function for cropping and padding the image based on the box prompt. The feature corresponding to the region of interest in the image can be expressed as:

$$h_{i,j} = f_{CLIP}\left(\text{crop}(x_{i,j}, p_{i,j})\right), \tag{10}$$

Next, we compute two statistics, the Gram matrix $G$ and the class prototype matrix $C$, defined as:

$$G = \sum_{i=1}^{T}\sum_{j=1}^{N_i} h_{i,j} \otimes h_{i,j} = \sum_{j=1}^{N_1} h_{1,j} \otimes h_{1,j} + \sum_{j=1}^{N_2} h_{2,j} \otimes h_{2,j} + \cdots + \sum_{j=1}^{N_T} h_{T,j} \otimes h_{T,j}, \tag{11}$$

$$C = \sum_{i=1}^{T}\sum_{j=1}^{N_i} h_{i,j} \otimes y_{i,j} = \sum_{j=1}^{N_1} h_{1,j} \otimes y_{1,j} + \sum_{j=1}^{N_2} h_{2,j} \otimes y_{2,j} + \cdots + \sum_{j=1}^{N_T} h_{T,j} \otimes y_{T,j}, \tag{12}$$

Notably, these statistics are computed as cumulative sums over samples, meaning they can be incrementally updated during the continual learning process. This eliminates the need to store image samples or prompts from previous tasks; only the accumulated results of the statistics from prior tasks need to be stored.

Using the least squares framework, the ridge regression classifier parameters can then be computed as:

$$W_o = \arg\min_{W}\left(\left\|Y_{\text{train}}^T - W^T H\right\|_2^2 + \lambda \|W\|_2^2\right) \tag{13}$$

which has the closed-form solution:

$$W_o = (HH^T + \lambda I)^{-1} H Y_{\text{train}} = (G + \lambda I)^{-1} C, \tag{14}$$

Here, $H$ is the matrix whose columns are all $h_{i,j}$, and $Y_{\text{train}}$ is the matrix whose columns are all $y_{i,j}$.

Subsequently, for any test image $x$, and its corresponding box prompt $p$, we compute the score vector $s$ as:

$$s = f(\text{crop}(x, p))W_o, \tag{15}$$

The $i$-th element of $s$, $s_i$, represents the score for the $i$-th class. Finally, the predicted class label $\hat{y}$ is determined as:

$$\hat{y} = \arg\max_i s_i, \tag{16}$$

Let $\tau_{\hat{y}}$ represent the task index corresponding to $\hat{y}$. The final mask prediction is then given by:

$$\hat{m} = f_{dec}\left(f_{enc}(x; \mathcal{I}), f_{pr}(p); \mathcal{E}_{\tau_{\hat{y}}}\right) \tag{17}$$

This incremental ridge regression approach ensures that EvoSAM can efficiently classify tasks without requiring access to previous task data while maintaining strong performance.

The above strategy ensures efficient task adaptation in the mask decoder while preserving the generalization capacity of the frozen image encoder, effectively mitigating forgetting and improving segmentation performance across diverse tasks.

## Details of Baselines

In our experiments, we implemented four baseline models for SAM-based continual learning: SAM-Seq FT, SAM-ER (25%), SAM-Distill, and SAM-EWC. SAM-Seq FT performs sequential fine-tuning across multiple tasks. SAM-ER (25%) retains 25% of the training samples from each previous task and fine-tunes the model on both the cached samples and the new task samples. While this approach can alleviate catastrophic forgetting, it often faces challenges in clinical scenarios due to storage limitations or restricted access to sensitive data. SAM-Distill introduces a distillation constraint by incorporating a copy of the model trained on previous tasks during fine-tuning of the new task. SAM-EWC estimates the importance of model parameters using the Fisher information matrix and uses these importance scores as regularization weights to constrain updates to parameters critical for previous tasks during fine-tuning on the new task. Both SAM-Distill and SAM-EWC aim to mitigate forgetting by constraining model updates during the learning process; however, these constraints can compromise the acquisition of knowledge for new tasks. Furthermore, none of these methods address the overfitting issue in SAM-based continual learning. To establish an upper bound for continual learning performance, we additionally fine-tuned a model using LoRA on the full dataset across all tasks. In addition, we extended these strategies to MedSAM by reinitializing the model parameters prior to continual learning. Following MedSAM, for all baseline methods, we fine-tuned the image encoder and mask decoder while keeping the prompt encoder frozen.

## Implementation Details

In our implementation, we simulate the user-provided box prompts by randomly expanding the boundary of the ground-truth mask up to 100 pixels for blood vessel segmentation, and 20 pixels for multi-site prostate segmentation. We employ diverse data augmentation techniques to enrich the training data diversity, including random cropping, horizontal/vertical flipping, 90° or 270° rotation, and color adjustments (brightness, contrast, saturation within [0.8, 1.2], and hue within [-0.1, 0.1]). All images are padded and resized to a resolution of $1024 \times 1024$ before input to the SAM image encoder.

We use the Adam optimizer[45] for training, with a learning rate of 1e-3 for LoRA fine-tuning and 1e-4 for full-parameter fine-tuning. The loss function consists of a combination of cross-entropy loss and dice loss, with equal weights (1:1). We set the EWC regularization weight to 100 and the Distill weight to 1e-5.

## Evaluation Metrics

Following prior studies, we measure the performance of our EvoSAM using the average mDice and average Forgetting across all tasks of the whole continual learning process. Specifically, we denote EvoSAM continually learned from T tasks as $M_{t=1}^T$. For an image $x_{i,j}$ from the $i$-th task, $M_t$ outputs a segmentation mask $\hat{m}_{i,j}^t$. The dice score between $\hat{m}_{i,j}^t$ and ground-truth mask $m_{i,j}$ can be calculated as:

$$\text{dice}_t^{i,j} = \frac{2|\hat{m}_{i,j}^t \cap m_{i,j}|}{|\hat{m}_{i,j}^t| + |m_{i,j}|} \tag{19}$$

Then, the segmentation accuracy for all the images of the $i$-th task can be computed as the mean dice across all images:

$$\text{mDice}_t^i = \frac{1}{N_i} \sum_{j=1}^{N_i} \text{dice}_t^{i,j} \tag{20}$$

where $N_i$ denotes the number of images in the $i$-th task. Finally, the segmentation accuracy of model $M_t$ for all tasks can be calculated as the average of mDice across all tasks:

$$\text{avg\_mDice}_t = \frac{1}{T} \sum_{i=1}^T \text{mDice}_t^i \tag{21}$$

Similarly, we can measure the catastrophic forgetting of model $M_t$ by computing the average Forgetting as follows:

$$\text{avg\_Forgetting} = \frac{1}{T-1} \sum_{t=2}^T \frac{1}{t-1} \sum_{i=1}^{t-1} \left(\text{mDice}_{t-1}^i - \text{mDice}_t^i\right) \tag{22}$$